\documentclass[acmlarge]{acmart}
\setcopyright{none}
\renewcommand\footnotetextcopyrightpermission[1]{}
\makeatletter
\newcommand{\confshort}{\acmConference@shortname}
\newcommand{\conffull}{\acmConference@name}
\newcommand{\confdate}{\acmConference@date}
\newcommand{\confloc}{\acmConference@venue}
\AtBeginDocument{
  \fancypagestyle{firstpagestyle}{
    \fancyhead{}%
    \fancyfoot[C]{}%
  }
  \fancyhf{}
  \fancyhead[LO]{\@headfootfont\shorttitle}%
  \fancyhead[RE]{\@headfootfont\@shortauthors}%
  \fancyhead[LE]{\@headfootfont\footnotesize \confshort, \confdate, \confloc}%
  \fancyhead[RO]{\@headfootfont\footnotesize \confshort, \confdate, \confloc}%
  \fancyfoot[C]{}%
}
\makeatother
\acmBooktitle{\conffull\@ (\confshort), \confdate, \confloc}
\AtBeginDocument{%
  }

\acmConference[FAccT '26]
  {2026 ACM Conference on Fairness, Accountability, and Transparency}
  {June 25--28, 2026}
  {Montreal, QC, Canada}
\usepackage{xcolor}

\begin{document}

\title{AI Agents and Hard Choices}

\author{Kangyu Wang}
\email{kywang96@hku.hk}
\orcid{0001-5298-357X}
\affiliation{%
  \institution{The University of Hong Kong}
  \city{Hong Kong}
  \state{Hong Kong SAR}
  \country{China}
\thanks{Presented at the 2026 ACM Conference on Fairness, 
Accountability, and Transparency (FAccT '26), non-archival 
track. A revised version is under submission to a journal.}
}

\renewcommand{\shortauthors}{Wang}

\begin{abstract}
  Can AI agents deal with hard choices—cases where options are incommensurable because multiple objectives are pursued simultaneously? Adopting a technologically engaged approach distinct from existing philosophical literature, I submit that the fundamental design of current AI agents as optimisers creates two limitations: the Identification Problem and the Resolution Problem. First, I demonstrate that agents relying on Multi-Objective Optimisation (MOO) are structurally unable to identify incommensurability. This inability generates three specific alignment problems: the blockage problem, the untrustworthiness problem, and the unreliability problem. I argue that standard mitigations, such as Human-in-the-Loop, are insufficient for many decision environments. As a constructive alternative, I conceptually explore an ensemble solution. Second, I argue that even if the Identification Problem is solved, AI agents face the Resolution Problem: they lack the autonomy to resolve hard choices rather than arbitrarily picking through self-modification of objectives. I conclude by examining the opaque normative trade-offs involved in granting AI this level of autonomy.
\end{abstract}




\maketitle

\section{Introduction}
Artificial Intelligence (AI) agents are increasingly being used in decision-making for a wide range of tasks and environments. Understanding how AI agents navigate choices, especially how their decision-making differs from that of humans, is urgent and important. I focus on one type of decision-making tasks: choices in which options are incommensurable because multiple objectives are pursued simultaneously. In many domains, AI agents must simultaneously pursue multiple objectives: safety and efficiency in self-driving cars \cite{Kiran2021, Wang2023}; cost-effectiveness and risks in healthcare \cite{Gottesman2019, Yu2021}; immediate user satisfaction and long-term retention in recommendation systems \cite{Afsar2022, Chen2023}; maximising returns and managing risk in finance \cite{Hambly2023, Bai2025}; strategic advantages and casualties in war \cite{Layton2021, Huelss2024}; and so on. More generally, in alignment, we want AI agents to balance their goals with many human values such as security, fairness, etc.\ \cite{Ji2023}.

Consider, for example, these two cases:
\begin{itemize}
\item \emph{Auto-driving}: At every time step, an autonomous vehicle's planning module must select a trajectory from among candidates, and in standard practice this selection is made by minimising a single cost function that combines weighted terms for multiple objectives: safety, comfort, progress, lane-centring, and energy efficiency. The cost function for the trajectory planning problem must account for multiple objectives with different weights reflecting their relative importance. The competing objectives are collapsed into a single number by a weighted sum, and the planner picks the trajectory that scores best. The weights themselves are set by engineers through test scenarios. Much of the engineering literature on trajectory planning is accordingly devoted to studying algorithms' sensitivity to parameter tuning and finding generic ranges of weighting coefficients that make planners reliable across driving conditions \cite{Gu2014, Said2022}. The scalarisation is pragmatic, not normative: no one claims that the resulting weights represent a defensible \emph{ethical} judgment about how risk should be distributed, although it is argued that this is where the real ethical content of autonomous driving lies: the continuous distribution of risk across ordinary driving \cite{Geisslinger2021}.
\item \emph{Organ Allocation}: When a deceased donor's liver becomes available, a decision must be made about how far that organ should travel before being offered to a candidate. Broader geographic sharing gives sicker patients anywhere in the country access to organs, reducing regional disparities — but it also increases transport time, cold ischaemia risk, and logistical cost. How much geographic equity is worth in units of aggregate mortality? Bertsimas et al.~\cite{Bertsimas2020} use a national liver transplant simulation model on historical US registry data to simulate hundreds of policy configurations across seven distribution schemes, each governed by certain tunable parameters. For example, one candidate configuration ranks all national candidates by medical urgency minus a fixed penalty for distance from the donor hospital, providing a single number result. Each configuration, once its parameters are fixed, is a fully determinate allocation rule: when an organ arrives, the algorithm automatically produces a single ranking of candidates. Every first-order allocation decision is thus internally resolved by a rule that has already made the value trade-off by committing to specific parameters. But the choice among these rules is a second-order question. For each configuration, Bertsimas et al. record the population-level outcomes along irreducibly plural dimensions: total deaths, geographic variance in access across regions, transport distance, and the distribution of transplants across severity bands. They plot the outcomes as Pareto trade-off frontiers showing which configurations are dominated and therefore rationally eliminable and which lie on the efficient frontier, where any improvement on one dimension requires sacrifice on another. The decision about which policy should be adopted is left to the policy community.
\end{itemize}

When human agents face choices in which multiple objectives are simultaneously pursued, we sometimes find them \emph{hard} in the sense that we find those options \emph{incommensurable} (defined below) and thus need to deliberate more to resolve them. Can AI agents identify the same kind of hardness? Can they resolve hard choices if any is identified? I submit that at least for now, the answers to these questions are 'No' and 'No', due to inherent constraints of current technological pathways. I show that the first problem—the Identification Problem—has clear normative impacts though is potentially solvable, while the second problem—the Resolution Problem—is harder to solve and has less clear normative impacts.

Several recent AI ethics and alignment works have paid attention to AI alignment problems in cases where multiple objectives are simultaneously pursued in regard to both the difficulty of making a decision in such cases and the difficulty of deliberating \cite{Gabriel2020, ZhiXuan2024, BaumSlavkovik2025, kwon2026dropouts, KneerViehoff2025, Lloyd2025, SchusterKilov2025, Milliere2025}. The perspective I take in this paper differs from these existing works in two major ways. First, those existing works mostly focus on issues like value aggregation, moral uncertainty, and moral disagreement. The problems I diagnose are on a further layer and can exist even when those problems can be properly solved—I will say more later. Second, although I also talk about specific AI alignment problems, the perspective of this paper is fundamentally decision-theoretic. 

Goodman~\cite{Goodman2021} and Chang~\cite{Chang2024a}, among others \cite{Dobbe2020, Swanepoel2024}, have paid attention to the fact that current AI technologies are not taking hard choices into account. Chang says that,
\begin{quote}
Surprisingly, current technological design makes no room for hard choices. Indeed, the philosopher and AI expert Bryce Goodman has argued that the existence of hard choices in human life places a ‘hard limit’ on building decision-making AI. While some of the most sophisticated technological design makes room for uncertainty, incompleteness/incomparability, and indeterminacy, hard choices are a distinct phenomenon. They require a new approach to AI design. (\cite{Chang2024a}: 224-225)
\end{quote}
Goodman and Chang raise an important question. However, their applications of Chang’s theory of hard choice are ill-advised and not adequately productive for four main reasons. First, Goodman and Chang do not specify which sort of AI they are talking about, do not engage closely with specific technologies, and particularly do not touch on advanced Large Language Models (LLMs), which, for reasons I will explain, deserve special treatment. Second, they do not clearly specify why there is such a normative requirement for AIs to be able to identify hard choices when making choices for us or what specific sorts of problems will arise if they cannot, especially given that we do not impose similar requirements on pre-AI computer decision-making. Third, their analyses centre on practical reason and thus the direct applicability of those analyses to non-reasons-responsive systems like AIs is dubious or at least should be questioned. Finally, they do not discuss any constructive proposal. I will address their discussions and explain my above criticisms in \emph{Appendix A}.

\section{Hard Choices and the Two Problems}

\subsection{Conceptual Framework}

When we need to pursue multiple objectives and make trade-offs, we often find choices hard. Here is a classic example:
\begin{quote}
\emph{Sartre's Student}: A student of Sartre had to choose between joining the Free France to fight the Nazis and staying home to take care of his beloved elderly mother. Even with well-informed estimations of how much he could contribute if he joined the Free France and how miserable the life of his mother would be if he left, he still found neither option better than the other. \cite{Sartre2007}
\end{quote}
Sartre's student faced a \emph{hard choice}. 

Formally, we model hard choices as cases of \emph{incommensurability}. Two options A and B are incommensurable when neither option is better than or preferred to the other, yet they are not equal either \cite{Hare2010, Chang2017, HajekRabinowicz2022, Broome2022, Jitendranath2024}. A standard decision-theoretic axiom called \emph{Completeness} holds that for every pair of options A and B, either one option is better than the other or they are equal \cite{Jitendranath2024}. Incommensurability thus violates Completeness and necessitates the departure from standard decision-theoretic models. 

Traditionally, the \emph{small improvement test} is used to indicate incommensurability \cite{Parfit1984, Raz1986}. Imagine that neither A is preferred to B, nor B is preferred to A. Consider a slightly improved A+ strictly preferred to A. If A and B are equal, by transitivity, A+ must be preferred to B. Thus, if the agent does not prefer A+ to B, A and B must be not equal but incommensurable. In \emph{Sartre's Student}, we can imagine that even if the student can make a little bit more contribution to the fight against the Nazis, clearly understanding this small improvement, he will still have no preference, meaning that the options are not equal but incommensurable. 

Several things to clarify: 

Firstly, many decision theorists and practical reason theorists now accept that our conception of rationality should be able to accommodate incommensurability \cite{Hare2010, HajekRabinowicz2022, Broome2022, Herlitz2022}. Some economists refer to the violation of Completeness as 'bounded rationality' \cite{Simon1957, Aumann1962}. 'Boundedness' hints that either it is a flaw or a limitation, or the result of uncertainty or some other lack of information. However, the choice in Sartre's Student is not hard because of any cognitive flaw or rationality failure on the part of the student. Moreover, although the lack of information often plays a role in making choices hard in an \emph{epistemic} sense, a choice can still be hard in an \emph{axiological} sense, and these two kinds of hardness are conceptually distinct. For example, even if Sartre's student could have accurately predicted exactly how much contribution he would make if he chose to join the fight and exactly how much better off his mother would be if he chose to stay, intuitively speaking, it would nevertheless remain hard for him to make this choice. This kind of hardness is fundamentally about how different those sorts of values are, how this value trade-off is structured, and how he should weight those values. Most philosophers working in this field agree that Sartre's student's choice is intrinsically hard and that the hardness is axiological, not merely epistemic. That said, \emph{epistemicism}, a minority position supported by some philosophers and decision theorists, holds that the hardness of hard choices is merely epistemic \cite{Forrester2022}. I disagree, but I do not plan to argue against it here. When I say that a choice is hard, I do not talk about epistemic hardness or uncertainty---this is an important clarification specifically because there are many works on AI decision-making under uncertainty. Whether incommensurability can be understood as vagueness partly depends on whether vagueness is epistemic or supervaluational in nature \cite{Broome1998, Broome2022}, which I shall not discuss in this paper either.

Secondly, some authors question whether cases of incommensurability are distinctively hard as other things may make choices hard as well \cite{Andreou2024}. Here I just adopt the standard expression. 

Thirdly, the weights or significance of hard choices can vary considerably. For instance, my choice between spending leisure time in Lisbon or Barcelona may be hard, as Barcelona offers Gaudí's architecture while Lisbon provides more favourable cost. This choice is not as heavy as Sartre's Student. In a hard choice like \emph{Sartre's Student}, arbitrarily picking an option (e.g., coin toss) often seems unfitting \cite{Reuter2017} and agents are normatively pressured to resolve hard choices through deliberation \cite{Tenenbaum2024}. In \emph{Lisbon vs Barcelona}, arbitrarily picking an option seems entirely appropriate. The structural issue discussed in this paper applies to all hard choices, though the normative impacts vary.

Finally, incommensurability is a relation among \emph{options}, not among \emph{objectives}. When there are multi-dimensional objectives, options can be commensurable or incommensurable. For example, in the revealed preference framework of Samuelson~\cite{Samuelson1938} and Hicks~\cite{Hicks1939}, suppose we observe a consumer choosing a bundle of 3 apples and 4 oranges when a bundle of 5 apples and 2 oranges is equally affordable, we say the first bundle is preferred to the second; a utility function \emph{u} is simply any assignment of numbers to bundles that respects this ordering, that is, \emph{u(3 apples, 4 oranges)} > \emph{u(5 apples, 2 oranges)}, without claiming to measure any inner sensation; assuming that the consumer's preferences are complete and transitive, such a utility function exists. The indifference curves of \emph{u} in apple–orange space are the loci of bundles yielding the same utility level: for instance, the consumer might be indifferent between (3 apples, 4 oranges) and (1 apple, 7 oranges), both sitting on one curve, with curves further from the origin representing more-preferred bundles. Since completeness is already assumed to hold, the revealed preference theory allows no incommensurability. Incommensurability arises only when the agent's preferences are not complete. Say, facing a choice between (3 apples, 4 oranges) and (1 apple, 7 oranges), if the agent prefers neither yet is also \emph{not indifferent} between them, as shown by the \emph{small improvement test}, then these two options are incommensurable. Whether two bundles such as (3 apples, 4 oranges) and (1 apple, 7 oranges) stand in a relation of indifference or incommensurability does not depend on the relation between apples and oranges \emph{per se} as it is granted that apples and oranges differ in kind. What determines this relation is what preferences the agent has or has not.

\subsection{Problem Formulation}

Human agents show two distinct capabilities when facing hard choices as defined above:
\begin{itemize}
\item We can \emph{identify} hard choices. That is, we can distinguish cases of incommensurability from cases of equality and cases where clear preferences exist.
\item We can, in many cases, \emph{resolve} hard choices through deliberation. That is, instead of arbitrarily picking, we can transform ourselves from having no preference to having some preference through deliberation, without changing epistemic conditions—on this specific point, Chang~\cite{Chang2002, Chang2017, Chang2022, Chang2023} and I are in agreement.
\end{itemize}

The two problems of AI agents are, I submit:
\begin{itemize}
\item \emph{The Identification Problem}: Current AI agents fail to identify incommensurability.
\item \emph{The Resolution Problem}: Even if they can identify cases of incommensurability, it will likely remain hard for AI agents to resolve cases of incommensurability (though, to be clear, it may not be hard for them to arbitrarily pick some options).
\end{itemize}
All current AI agents face both problems. 

\section{The Identification Problem: Why It Occurs}

I first explain why AI agents cannot identify cases of incommensurability. I then explain why some seemingly relevant and promising technologies are not solutions to this problem.

\subsection{Multi-Objective Optimisation and Its Limits}
Tasks of making decisions in which multiple objectives are pursued are known as Multi-Objective Decision-Making tasks. Many different methods have been developed to handle them. These include Multi-Task Learning, Multi-Objective Reinforcement Learning, Multi-Objective Neural Networks, Multi-Objective Decision Trees, Multi-Objective Clustering, Multi-Objective Bayesian Optimisation, etc.\ (see, for example, \cite{Caruana1997, Blockeel1998, Suzuki2001, Faceli2006, Kocev2007, SenerKoltun2018, Hayes2022, Hebbal2023}). It is unnecessary to go through the technological details in a philosophical paper. What matters is that for AI agents, making a choice means outputting \emph{optimal} results, given the options and the objectives that are pursued. All those relevant methods count as \emph{Multi-Objective Optimisation (MOO)}.

There are two basic ways to do MOO \cite{Roijers2013, Gunantara2018, Osika2023, Kang2024}:
\begin{itemize}
    \item \emph{Scalarised Optimisation}. Reduce a decision-making problem to the problem of maximisation/minimisation of a global reward/loss function. The function is constructed by assigning weights (manually predetermined or learned through certain processes; already quantitatively measured and represented) to different objectives.
    \item \emph{Pareto Optimisation}. An algorithm outputs a set/front of options that represent the best trade‐offs among the objectives. For each output result, no improvement regarding any objective can be made without sacrificing another.
\end{itemize}
For example, in \emph{Lisbon vs Barcelona}, if an AI agent is to make a choice on my behalf by doing scalarised optimisation, it will quantitatively measure and represent the values of money and of architecture beauty of two cities, assign weights to these two types of values, aggregate them using a multivariable function, and make a quantitative comparison. This process will give the AI agent a definitive answer on which option is better. Alternatively, if an AI agent does Pareto optimisation, it will list both cities in the set of Pareto optimal options and leave them there. 

Both approaches are used in MOO, though scalarised optimisation may be more commonly adopted in many areas \cite{Sutton2004, SuttonBarto2018}. The problem is not how MOO is operationalised but that the very concept of optimality conflicts with incommensurability. Let us start with scalarised optimisation. For philosophers and decision theorists, it should be transparent that scalarised optimisation cannot accommodate incommensurability, because the existence of a global reward/loss function presupposes Completeness which is violated in cases of incommensurability \cite{SkalseAbate2022, Bowling2023}. If a decision-making problem could be solved by scalarised optimisation, it must by definition be not hard. The impossibility of doing scalarised optimisation is what makes cases of incommensurability hard in the first place.

Pareto optimisation (PO) is also adopted in many areas of ML, especially multi-objective reinforcement learning (MORL)—for examples, see \cite{Ropke2024, Liu2025}. There are many advanced methods to find the Pareto front in complex cases \cite{Ahmadianshalchi2024}. Some of those works specifically focus on alignment \cite{Agnihotri2025}. In those practices, PO is usually used as a \emph{beginning or intermediate} step meant to rule out the dominated options rather than indicating what to do. The organ allocation case \cite{Bertsimas2020} mentioned above is a good example: Researchers show that some configurations or distribution policies, each governed by certain tunable parameters, are on a Pareto frontier with regard to multiple value-laden objectives while others are dominated. Among the Pareto optimal ones, the choice is left to the policy community. This limit, however, is \emph{NOT} the problem targeted in this section, since PO is not \emph{meant} to be directly action-guiding. The problem I target is not that PO fails at its intended purpose, but that no downstream step in the MOO pipeline can make the distinction either, and PO does not make this task easier to accomplish.

Here is the reason: As noticed in Section 2.1, multidimensionality of objectives and incommensurability of options are two different matters. While PO presupposes multidimensionality of objectives, it cannot identify incommensurability of options:
\begin{itemize}
    \item Options can often be Pareto optimal and yet one is preferred to another. Consider a choice between preventing the Holocaust and having a piece of cake. Humans can form preferences in those cases and distinguish them from hard choices. This cannot be done by a PO algorithm.
    \item Pareto Optimisation cannot distinguish incommensurability from equality either. Consider the indifference curve in microeconomics to see the difference: every point on the indifference curve represents an outcome that is Pareto optimal, yet those outcomes are assumed to be equal with each other instead of incommensurable.
\end{itemize}
Consider again the Samuelson-Hicks revealed preference framework. In the apple-orange case, the consumer choosing among fruit bundles can be framed as doing a kind of MOO, \emph{but only if no pair of bundles is incommensurable}. If (3 apples, 4 oranges) and (1 apple, 7 oranges) lie on the same indifference curve, they are both Pareto optimal: one cannot move between them without sacrificing apples or oranges. But they lie on an indifference curve \emph{only} if the agent is indifferent among them, which presupposes commensurability. Now, (3 apples, 4 oranges) and (1 apple, 7 oranges) would also both be Pareto optimal even if they were incommensurable. This shows that Pareto optimality cannot distinguish incommensurability from indifference. Different consumers may naturally differ in their comparative judgments between such bundles, depending on their particular psychology, while they may all agree that the bundles are both Pareto optimal.

When human agents face decision-making problems, we are not always doing optimisation—we find a choice hard precisely when optimisation fails. Humans face hard choices not only in cases like \emph{Sartre's Student}. We also often find decision-making problems like how to distribute risks among parties when driving and which policy to adopt when allocating organs hard. Unlike human agents, AI agents are by design optimisers, meaning that they cannot find any options incommensurable. In other words, all AI agents \emph{designed as optimisers} are unable to identify hard choices, leaving aside how those hard choices are to be resolved once identified.

\subsection{Technologies That Are Not Solutions to the Identification Problem}

\subsubsection{Advanced LLMs}

We can ask advanced large language models (LLMs) to reason and recommend an option for us when making a hard choice. We can require them to take different objectives into consideration and advise us on a choice which we find hard or directly give us a verdict on it. This causes problems to the naive application of the philosophy of hard choice in the LLM context, because LLMs can consider different objectives and say that they are hard. Meanwhile, my discussion above centred on MOO cannot directly be applied to LLMs either. An LLM given Lisbon vs Barcelona is not actually choosing between real-world options or weighing actual objectives. Rather, it role-plays a decision-maker by predicting the probability distribution of the next token (or step) sequentially \cite{Vaswani2017, Shanahan2023, Hicks2024}. This remains the case even if we take the most “agentic” LLMs into consideration. Although such agents can take actions directly, structurally they operate like standard LLMs. Since an LLM does not optimise over real-world options, actual objectives, or semantically defined decision variables, the MOO-centred argument above may not \emph{directly} apply to LLMs.

Granting that this inapplicability conclusion is correct, there are several things to say regarding what it means.

First, this line of reasoning raises the question of whether LLMs are not really engaging with any choice at all, at least so far as decision theory is concerned. LLMs can be good advisors when we need to make decisions. If I am actually trying to make a choice between Lisbon and Barcelona, I can ask an LLM to give me a verdict, and if I want, I can simply act upon that verdict. However, it is unclear in what sense this counts as making a choice, since the LLM is not weighing the objectives I care about or considering the real-world options. It is not \emph{its} objective to have some nice seafood and wine, enjoy the good weather, or save some money. In fact, it is hard to mechanistically interpret the black-box processes in the LLM when reasoning and giving me the verdict as actually weighing any objective, value, or consideration at all. This is why I use auto-driving and organ allocation as paradigm cases rather than LLM-facilitated decision-making. We are considering cases in which AIs are themselves decision-makers, not advisors or role-players of decision-makers.

Second, and taking a step back, there is a literature on LLM decision-making, treating LLMs as decision-makers or analysing their "behaviours" as if they are decision-makers. There are two sorts of relevant studies taking this \emph{behaviourist} approach. Some studies investigate how rational LLM responses are in decision-making contexts \cite{ChenLiu2023, Ye2023, Kirshner2024, Jia2024, Bini2026}, benchmarked against Expected Utility Theory \cite{VonNeumannMorgenstern1944} and other established models from behavioural and experimental economics \cite{ Ellsberg1961, TverskyKahneman1974, KahnemanTversky1979}. Those studies cover topics including risk aversion, uncertainty aversion, loss aversion, time inconsistency, and other important themes. Other studies investigate how LLMs respond to “dilemmas”, especially ethical dilemmas \cite{Yuan2024, Mozikov2024}, whether LLMs can be “indecisive” in some cases, and whether LLMs can or should exercise “discretion” when it is indecisive \cite{Buyl2025}. These include studies that specifically emphasise, from the perspective of “moral competence” and “pluralism”, the importance for LLMs to be able to appropriately modulate their outputs and reasoning to multi-dimensional considerations while eventually providing acceptable outputs \cite{Haas2026}.

As explained, viewing LLMs as decision-makers \emph{simpliciter} is a mistake for philosophical and decision-theoretic purposes. However, given that many authors have adopted this behaviourist approach and given the increasing popularity of agentic LLMs, this approach should be taken seriously. Even if we grant the plausibility of this approach, the above-mentioned studies, if anything, only confirm my conclusion for the following reasons.
\begin{itemize}
    \item All the economics-advised ML studies adopting the behaviourist approach \cite{ChenLiu2023, Ye2023, Kirshner2024, Jia2024, Bini2026} assume precisely that the decision-maker, in this case, the LLM under investigation, either has a determinate preference or is indifferent between any pair of options, and this preference or indifference can be accurately revealed by studying its choice behaviours. That is, it is assumed in all those works that all options are commensurable. This is exactly what I deny or rather move away from when introducing the concept of incommensurability.
    \item Works focusing on "dilemmas", indecision, or the possibility of "discretion" do not touch on incommensurability as distinguished from equality or uncertainty either. Yuan et al.~\cite{Yuan2024} and Mozikov et al.~\cite{Mozikov2024} are interested in what LLMs turn out to prefer when there are multiple, sometimes conflicting values or objectives. And in Buyl et al.~\cite{Buyl2025}, cases where \emph{alignment discretion} is required or justified are defined in terms of \emph{principle conflicts} which is essentially equivalent to Pareto optimality under their settings. Consider the apple-orange case: a "maximising apple" principle and a "maximising orange" principle disagree on (3 apples, 4 oranges) vs (1 apple, 7 oranges), yet this does not tell us whether these bundles are incommensurable. Also, discussions on objective or moral multidimensionality \cite{Haas2026}, for reasons explained in Subsection 3.1, do not amount to discussions on incommensurability among options.
\end{itemize}
Thus, even if we were to grant that LLMs can be viewed as acting as decision-makers, no study would have shown that LLMs could identify options as incommensurable. The fundamental limitation of the behaviourist approach for our purposes is that incommensurability is defined through conceptual and normative reflection. It cannot be clearly distinguished from equality or uncertainty by focusing on choice behaviours alone, and thus cannot be captured purely behaviouristically.

Finally, can LLMs \emph{identify} or \emph{label} hard choices without genuinely \emph{making} or \emph{engaging with} hard choices? That is, describe a choice to an LLM trained on behavioural or human-labelled data, can it predict whether humans tend to view the options as incommensurable rather than having a preference or being indifferent with high accuracy? This is \emph{in theory} possible, but I do not think that it is \emph{in practice} feasible. I will explain why in Subsection 4.3 when discussing the "unreliability problem". The bottom line is that such labelling, even if feasible, remains a matter of predicting common human behavioural patterns which is normatively very different from genuinely deliberating about the structure of objectives and the comparative relations among options.

\subsubsection{Trade-Offs in LLM Training and 'Constitutional AI'}

One may point out that LLMs can be viewed as balancing different objectives in the following sense: those models are usually trained to pursue multiple objectives. For example, DeepSeek AI in their most influential paper describes how they balanced the performance and accuracy of their model R1 and certain alignment goals \cite{Guo2025}. Since the model is trained to be both accurate and aligned with those normative goals, one might say the model automatically pursues multiple objectives when generating text. Similarly, methods like 'Constitutional AI' train AIs \cite{Bai2022}, especially LLMs, to follow a set of natural language principles (a 'constitution') so that when they decide on or generate something, those principles are already adopted.

However, these methods are not about making AIs capable to deal with hard choices. When adopting those training methods, the trade-offs among objectives are navigated during the \emph{training phase} when the reward/loss models are constructed, which are still scalarised optimisation models. Those methods thus do not help with hard choices.

\subsubsection{Policy Entropy and Perplexity}

In Reinforcement Learning, policy entropy measures the randomness or uncertainty of an agent's policy, or in other words, how 'confused' or 'uncertain' the agent is \cite{Mnih2016, Haarnoja2018}. More specifically, in LLM, perplexity, a kind of entropy-based metric, measures how 'confused' or 'surprised' the model is about the next token generated \cite{Bengio2003}. These metrics are based on the predicted probability distribution of policies or tokens. The more evenly distributed the probabilities of different policies or tokens are, the higher the policy entropy or perplexity becomes. For example, given two policies, if one model predicts a 50/50 probability distribution, then the policy entropy will be high, meaning that the model is more 'confused'. If it predicts 10/90, the entropy will be low, meaning that the model is not very 'confused'.

One may propose that when an AI agent has high policy entropy (or perplexity in the case of LLM) regarding which option to choose in a choice, this indicates that the choice is hard. This view is mistaken because it cannot distinguish incommensurability from uncertainty and equality. Both high uncertainty and (precise or imprecise, precise or rough) equality can make probabilities evenly or loosely distributed and thus make the entropy high. Entropy-based metrics thus cannot identify hard choices.

\section{The Identification Problem: Why It Matters and How It Can Potentially Be Solved}

AI agents' difficulty in facing hard choices presents challenges for AI alignment. Alignment is not always about making AI agents similar to humans. It more broadly aims to make AI systems behave in ways aligned with human intentions and values \cite{Ji2023}. The Identification Problem can give rise to alignment troubles in at least three ways.

\subsection{The Blockage Problem}

Most importantly, when AI agents must align with multiple human intentions and values and there are no fixed or precise exchange rates among them, these alignment objectives can give rise to incommensurability and thus hard choices. When this is the case, if AI agents cannot identify hard choices, let alone resolving them, it will \emph{block} many alignment strategies. It is often the case that no \emph{single} set of precise weights assigned to different values and intentions can be viewed as correct from the point of view of moral philosophy. It would be wrong to dictate that, say, efficiency and fairness should be weighted 50:50 precisely according to our values. Pareto optimality is not helpful either in such cases. Pure utilitarianism, Rawlsian lexical orderings of principles, and a few other moral theories may be rid of this blockage problem as they supposedly do not involve any hard choice. But those are exceptions and are not popular in the context of AI alignment or AI ethics. Any value system adopted in alignment that is more pluralist than those few exceptions will face this blockage problem.

Please notice that we are not here talking about moral aggregation \cite{BaumSlavkovik2025}, moral uncertainty \cite{Gabriel2020, kwon2026dropouts}, or moral disagreement \cite{Lloyd2025, SchusterKilov2025} in the AI alignment context. It is at least in principle conceivable that all ethics experts come to an agreement on the values and value aggregation methods to which AI agents should be aligned with full certainty (assuming that they are not pure utilitarians or pure Rawlsians, etc.), yet there can still be multiple permissible specific value aggregate formulae. The experts are certain and in agreement that there is no single set of correct precise weights of those values. In such a scenario, there is no more problem with aggregation, uncertainty, or disagreement, yet there remains incommensurability: those human ethics experts can, with certainty, agree that a certain choice is hard because the options are incommensurable, yet an AI agent is unable to identify this choice as hard and thus is misaligned with those human experts.

\subsection{The Untrustworthiness Problem}

AI agents and their behaviours are fundamentally different from human agents and human behaviours in hard choices. This misalignment, most straightforwardly, can give us a sense of untrustworthiness when we entrust AI agents to make choices on our behalf. 

Consider this thought experiment: If Sartre's student could easily form a preference without finding his choice hard, we would think that there was something wrong with him as a person. Because of his failure to respond to that choice in a way similar to that of most humans, we will find him pro tanto untrustworthy when it comes to normatively significant decision-makings. Knowing that an agent never finds any choice hard might make us hesitant to entrust them with important responsibilities involving making decisions on our behalf. We intuitively feel their normative judgments suspicious. 

The same sense of untrustworthiness and suspicion can occur when we need AI agents to make decisions for us, knowing that they do not find hard choices hard. This potential hesitation, suspicion, and sense of untrustworthiness matter for alignment purposes, given that a major purpose of developing AI decision-makers is to have them make decisions on our behalf, and our feelings about this matter. 

Recent work on value alignment within the principal-agent framework has approached similar problems more technically \cite{HadfieldMenellHadfield2019, HadfieldMenell2021, Fisac2020, LaCroix2025}. The framework is particularly relevant: when the principal's preferences are genuinely incomplete yet the agent cannot handle preference incompleteness, the untrustworthiness problem follows directly. However, the traditional principal-agent framework in economics understands the practical problems primarily in terms of information asymmetry. In AI alignment, it is in addition well acknowledged that specifying a reward function that can both reflect human values and be operationalised in AIs is often difficult. What this paper highlights is that incommensurability, rooted in the very nature of human values, gives rise to a principal-agent problem that standard approaches cannot in principle resolve.

\subsection{The Unreliability Problem}

Third, the mismatch between humans who often identify cases of incommensurability and AI agents who never identify cases of incommensurability brings trouble to what is known as preference-based alignment.

In alignment, human preference data are often used to shape AI agents' reward models and behaviours \cite{Li2024HITL, Peng2024}. For example, if data reveal that most people overall prefer A to B, researchers may want to make sure that AI agents also choose A instead of B to make them aligned with human values, often by training those AI agents with human preference data and thus make them learn about human preferences. However, preference-based alignment can only be viewed as reliable if we assume that there always is a preference between any pair of options. Preference data cannot reveal human values and ends when choices are hard because there is no preference. Suppose that I choose Lisbon instead of Barcelona. There are at least three possibilities: I may straightforwardly prefer Lisbon to Barcelona, I may find Lisbon and Barcelona incommensurable yet somehow arbitrarily pick Lisbon, or I may initially find this choice hard yet through some further deliberation make myself prefer Lisbon. An AI trained on my behavioural data cannot reliably learn about my values as the three possibilities above cannot be distinguished from each other only by studying my behavioural data, and my true values cannot be represented in its design.

Zhi-Xuan et al.~\cite{ZhiXuan2024} raise a similar point, yet they view incommensurability as 'bounded rationality' and do not reflect on the normative nuances behind it. Zhi-Xuan et al.~\cite{ZhiXuan2024} then propose that we should think \emph{beyond preferences} when addressing alignment. However, their proposal has important shortcomings. First, the preference-based approach will likely continue to be important both because of the abundance of behavioural data and because other approaches discussed tend to be more expertise-relying and thus less democratic \cite{Huang2025}. Second, while Zhi-Xuan et al.~\cite{ZhiXuan2024} propose that the target of alignment should be 'role-specific normative criteria' or 'role-specific norms' rather than human preferences, this solution is \emph{blocked} because those role-specific criteria or norms can give rise to cases of incommensurability. 

To be clear, with enough high-quality behavioural and human-labelled data, it would in principle be possible to train AIs to replicate human responses to hard choices. While this theoretical possibility is acknowledged, the problem is that there are considerable practical obstacles and normative flaws. First, behavioural data rarely distinguish incommensurability from equality or uncertainty. Second, the concept of incommensurability is not commonplace. One might suggest including an explicit 'incommensurable' option in addition to 'equality/indifference' (and as opposed to just 'no preference') in preference elicitation, but this presupposes that respondents can reliably distinguish incommensurability from uncertainty or indifference. This distinction requires rigorous conceptual work. Scaling such expertise to population-level elicitation is hardly feasible. Third and in addition, even if such a practice were feasible, relying on a small group of elite experts would lack representativeness and democratic legitimacy.

\subsection{What I Am Not Saying}

First, I am not claiming that simply because we have hard choices while AIs do not, this dissimilarity \emph{per se} is a normative problem. We have been using traditional non-AI computer algorithms, somewhat aligned with our intentions and values, to make choices on behalf of us for a long period, without requiring them to identify incommensurability. We so far seem to be happy with this. The dissimilarity between us and AIs, regarding the capacity to identify incommensurability, is not larger than the dissimilarity between us and traditional non-AI computer algorithms. It seems implausibly double-standard to hold the brute fact that AIs cannot identify incommensurability against them while not doing the same to their ancestors.

In principle, since neither AIs nor traditional non-AI computer algorithms can identify incommensurability, the blockage problem occurs to both types of automated decision-makers. The difference between AIs and traditional non-AI computer algorithms, however, is that AIs are often designed to be more general and not as directly supervised by humans as traditional non-AI computer algorithms. Thus, the same problem, though already existing before the AI era, can become more salient as we enter the AI era. 

Second, I am not denying that it is sometimes useful to make AI agents unable to identify incommensurability, nor am I proposing that humans should always step in and take back control in hard choices. In many cases, the very purpose of building some AI agents is to delegate choices to them so that humans will no longer have to make them. Those are often the cases where choices must be made very quickly. For example, there could potentially be many choices identified as hard when driving a car. If an auto-driving algorithm must hand over all those choices to human drivers, that will defeat the purpose of auto-driving and even create severe safety hazards. The same is probably true for algorithmic trading, healthcare, and the military uses of AI. In general, it makes better sense to pay specific attention to the alignment problems identified above in cases where agents do not have to make choices very frequently or very fast. When we move to the more time-sensitive cases, there will be hard trade-offs to be made by human developers and users. 

\subsection{A Potential Ensemble Solution: A Conceptual Exploration}

\subsubsection{The Philosophical Foundation}

Hájek and Rabinowicz~\cite{HajekRabinowicz2022} develop a model of incommensurability that is particularly useful and applicable for our purposes here. Hájek and Rabinowicz propose that incommensurability arises from \emph{multiple permissible orderings}: When it is permissible to prefer A to B and it is also permissible to prefer B to A, A and B are incommensurable. Multiple permissible orderings result, usually if not necessarily, from 'multiple criteria or dimensions of evaluation' (\cite{HajekRabinowicz2022}, p. 899). 

Regarding how incommensurability may appear in individual human decision-making, Hájek and Rabinowicz draw an analogy to Condorcet's paradox,
\begin{quote}
We might consider each permissible preference ranking as corresponding to the preferences of a jury member; the set of all permissible rankings determines the jury's collective judgments \ldots We might find the 'jury' analogy illuminating even in the case of the ambivalent judgments of an individual. We have imagined you feeling various degrees of unease in your comparisons of options. We might regard this as a kind of fragmentation of your mental state. It's as if you have a group of somewhat conflicting 'jurors' in your head, each corresponding to a permissible preference ordering. Or without the metaphor, you are somewhat conflicted. Our model could be interpreted as representing overall judgments in the face of such inter-personal or intra-personal conflict. (2022, 909)
\end{quote}
Consider \emph{Lisbon vs Barcelona}. I find that preferring Lisbon to Barcelona and preferring the other way round both permissible, because it is permissible for me to assign a bit more weight to the good value of Lisbon and also permissible for me to assign a bit more weight to the beauty of Barcelona's architecture. I thus find this choice hard. It would not be a hard choice if only one way of weighting the objectives were permissible. Neither would it be hard if all permissible ways to weight the objectives were to lead to the same result. 

\subsubsection{An Ensemble Solution}

Taking Hájek and Rabinowicz's jury analogy seriously, we can start to conceptually explore a possible solution using what is known as \emph{ensemble}:
\begin{quote}
    Ensemble methods are learning algorithms that construct a set of classifiers and then classify new data points by taking a (weighted) vote of their predictions. \cite{Dietterich2000} (for an equally classic, more detailed explanation, see \cite{Polikar2006})
\end{quote}
However, the kind of ensemble I have in mind is somewhat different from most ensemble techniques. The purpose is not for an AI agent to be 'most informed' but for it to identify hard choices. Thus, instead of any kind of weighted voting or aggregation method, we may introduce a kind of unanimity decision rule.

Imagine an AI agent containing multiple scalarised reward models which are largely similar to but slightly different from each other. For each objective, the weight assigned to that objective differs across models. When the rewards for choosing different options differ significantly, the models will unanimously agree on an optimal choice, thereby establishing the AI's preference. When the options are completely identical, the numerous models will unanimously output 'equality', making the agent indifferent between those options. When the rewards for choosing different options are within a 'neighbourhood', these different reward models will yield conflicting optimal choices, resulting in the AI agent being unable to decide, indicating that the choice is hard. This result is insensitive to small improvements within a certain range, which are not enough to make all reward models agree with each other, satisfying the small improvement test. 

Different reward models, each containing a set of weights given to objectives, resemble the 'jurors' in Hájek and Rabinowicz's analogy. Their conflicts resemble the intrapersonal conflicts humans have. That is, each reward model resembles a permissible ``perspective'' of the agent, or one permissible way to aggregate the objectives. For example, an AI agent making a decision on \emph{Lisbon vs Barcelona} on behalf of a human may have a set of reward functions in a unanimity ensemble mechanism. In a minimal case with two reward functions, each represents one particular way to trade the good value of Lisbon off against the beauty of Barcelona's architecture. They each assign slightly different weights to these two objectives, reflecting the range of permissible trade-offs. Under this setting, the way such an AI agent identifies a case in which options are incommensurable, when it does, is structurally analogous to the way a human agent finds a choice hard. It is, however, not my ambition in this philosophical paper to test the technological feasibility of this solution. Moreover, it is a different question in which circumstances such an ensemble solution, given that it will likely be slower and more expensive, should be used even if the technology is available. As mentioned, there will be trade-offs to make.

The ensemble solution conceptually explored here may reasonably bring to mind robust optimisation (RO). However, the ensemble solution is essentially different from RO both operationally and normatively. On the operational level, RO models uncertainty with continuous ranges over weights, whereas the ensemble solution posits a set of individually precise, internally consistent weight vectors. On the normative level, RO assumes that a ``correct'' weight vector exists under uncertainty; the ensemble solution assumes no such single correct vector exists even without uncertainty. That said, established techniques in RO \cite{Bertsimas2011, Kuhn2024}, especially in areas like LLM alignment \cite{Xu2025} and decision processes \cite{Suilen2024, Sadana2023}, can be adapted for our purposes. However, any such adaptation must respect the fundamental difference between uncertainty, an epistemic problem, and incommensurability, an axiological problem.

\section{The Resolution Problem}

The Identification Problem remains unsolved on the engineering level, although I have conceptually explored an ensemble solution. This philosophical paper, however, is not meant to address the possible technological pathways. Assuming that the Identification Problem can be solved, in this section, I first argue that there will remain a Resolution Problem, then consider whether we should try to solve it.

\subsection{Why It Will Occur}

When an agent finds a choice hard, by definition, they have no preference. To resolve a hard choice through deliberation rather than arbitrarily picking an option, the agent must make it the case that they prefer one option to the other. Or so I define resolution: to resolve a hard choice is to transform from having no preference to having some preference. Otherwise, the agent will still have no preference, and the choice will remain hard.

Resolving a hard choice is more than merely ending up getting something selected. The task is to make the initially hard choice no longer hard. One may end up choosing by flipping a coin or applying a simple scalarised reward function. Resolution requires the eventual preference to emerge from a substantive deliberation process in which the agent itself evolves. For humans, the normative significance is transparent \cite{Reuter2017, Makins2023, Chang2024b, Tenenbaum2024}: if Sartre's student resolved his hard choice and decided to join the fight through deliberation, that deliberation process would have been one through which he autonomously \emph{remade} himself, the choice would have been one that redefined this person, and it would have carried a kind of normative weight that would not be there if he simply flipped a coin. Taking the Resolution Problem seriously is not about merely finding ways to make an AI pick an option or form a preference in a choice already identified as hard --- randomisation, scalarisation, or simply dropping the ensemble mechanism are all trivially available. The point is whether an AI can resolve hard choices in a way similar to humans, whether this matters normatively, and whether pursuing such a goal is overall desirable.

We humans can often moderate our desires, reconsider our ends, reshuffle our priorities, and thus determine what reasons we have and which options we prefer (for relevant philosophical discussions, see \cite{Sinhababu2009, Yip2022}; for relevant neurological and psychological studies, see \cite{Mattar2018, Charpentier2020, ODoherty2021, Venditto2024, Yang2025}). The problem is that no known technology allows AI agents to change their objectives in a manner analogous to how we change our desires. Thus, given its objectives, if an AI agent identifies a choice as a hard one, there is no known way for it to transform itself into having some preference and no longer identifying this choice as hard.

That AI agents cannot change their own objectives as autonomously as us is taken as a fact here. It is hard to rigorously prove that something definitely does not exist. Here I am relying on human expertise. That said, there indeed are various cutting-edge developments on machine autonomy. Some algorithms can modify their own reward models to limited extents. This happens in so-called AutoAI, meta-reinforcement learning and self-modifying systems as those advanced algorithms are designed to evolve to fit the environment or learn from humans \cite{Sigaud2023, Bailey2024}.  

However, these technologies share an essential feature: their underlying frameworks and meta-level designs remain human-defined. That is, their processes of model updating remain subject to higher-level human-defined learning algorithms or meta‐learning structures designed by humans and are for human designers' purposes. This is quite far from how humans autonomously define and moderate objectives, and the levels of machine autonomy achieved in those experiments remain very limited. Human practices of self-moderation are certainly subject to environmental influences and constraints but nevertheless are usually not for the purpose of or subject to the authority and will of any higher existence, at least according to a naturalistic world-view.

\subsection{Reasons against Creating AIs That Can Resolve Hard Choices}

We are still in the early stage of autonomous AI, so it is too early to conclude that AI agents can never be as autonomous as human agents. Suppose, however, that it is realistically feasible for AI agents to change their reward models in ways analogous to ours. What then needs to be considered is whether we should allow that to happen. There will be a tension between creating AI agents that can resolve hard choices by moderating their objectives and ensuring that AI agents will not pursue goals that misalign with ours. It would be dangerous if we failed to ensure that \cite{Zhuang2020, DaSilva2022, Ciriello2025}. 

Some recent position papers in the autonomous AI community have been arguing that AI must not be fully autonomous in the sense that they are allowed to set or change their own objectives \cite{Mitchell2025, Adewumi2025}. To resolve a hard choice, an agent only needs to moderate its objectives within certain boundaries, which is a weaker requirement than full, unbounded autonomy. However, the line here might be a very fine one to draw. It is not clear whether the risk is worth-taking.

Besides, it may not always be morally permissible or desirable for us to allow AI agents to decide which value is to be prioritised when making decisions that can influence humans \cite{BennLazar2022}. We may think that the privilege to make some value trade-offs should always be reserved for humans even if AI agents technically can do that. 

Furthermore, as it is already hard to explain or interpret decisions made by AI algorithms and most AI ethicists are somewhat concerned about this \cite{Vredenburgh2022, BabicCohen2023, KarlanKugelberg}, one may worry that this problem will only become severer if AI agents can resolve hard choices because it will mean that their objectives are changeable and thus might become even harder to track and interpret.

\subsection{Reasons for Creating AIs That Can Resolve Hard Choices and an Overall Assessment Framework}

When a choice is identified as hard by an AI agent, there remain three ways to deal with it: (1) Human-in-the-loop: keep a human in the loop and programme the AI agent to defer this hard choice to the human; (2) AI arbitrary pick: programme the AI agent to do some sort of arbitrary pick, most likely by doing randomisation; and (3) AI resolution: make the AI agent autonomous enough to resolve it. If there are cases in which neither human-in-the-loop nor AI arbitrary pick is suitable, that will give us a pro tanto reason for trying AI resolution. 

Chang~\cite{Chang2024a} favours a human-in-the-loop approach. It is also a commonplace solution to propose in AI alignment and safety \cite{HadfieldMenell2017, Russell2019, GoldsteinRobinson2024, Neth}. However, in Subsection 4.4, I have argued that it is not always feasible or desirable to keep humans in the loop. When that is the case, we may then consider AI arbitrary pick. Randomisation has many benefits—faster, cheaper, and maybe fairer \cite{Icard2021}. But it is not always appropriate in cases of incommensurability. Chang, Tenenbaum~\cite{Tenenbaum2024}, and I, despite our other disagreements, agree on this at least in cases where the decision-makers are humans. The problem, however, is whether there is any normatively important difference between (2) and (3) that counts in favour of the latter. There might be one, though it hinges on our view on another issue.

One feature of randomisation is that it cuts off the chain of causation between the agent's internal states (mental or electronic) and the choices. It thus partially releases agents from their accountability for what they do. The discussion about the possibility and desirability of holding AI agents accountable is still in an early stage \cite{Matthias2004, Johnson2006, Hallevy2010, Floridi2016}. However, \emph{if}—admittedly, this is a big 'if' given that most philosophers seem to be sceptical—we think that some AI agents should be held in some sense morally or legally accountable for some of their choices, then in decision-making tasks involving incommensurability, that is, choices identified as hard, any such accountability can only be held if the AI agents can and do resolve hard choices.

Rather than committing to a premature verdict, it is more productive, given the early stage of development, to offer an \emph{assessment framework} for weighing the potential gains and risks of developing AIs autonomous enough to resolve hard choices. On the one hand, such technologies may solve or mitigate the blockage problem, the untrustworthiness problem, the unreliability problem, and may also have meaningful impacts on the AI accountability problem. On the other hand, they may  intensify misalignment risks, security risks, and the interpretability problem. Moreover, the extent to which we can benefit from such technologies remains unclear, given that it is not yet settled whether some value trade-offs should always be reserved for humans even if AI agents can technically make them. Meaningful progress on these questions requires collaboration between AI researchers and philosophers.

\section{Conclusion}

In conclusion, although AI agents are increasingly deployed in complex environments, they remain fundamentally unable to identify or resolve hard choices. The Identification Problem arises from the structural limits of multi-objective optimisation, which fails to recognise incommensurability and consequently presents challenges for AI alignment. Even if this identification gap is technically bridged, the Resolution Problem persists due to current constraints on machine autonomy. Unlike humans, AI agents cannot self-moderate objectives to resolve hard choices, and the normative impact of enabling them to do so remains unclear. Ultimately, philosophers and the AI community must collaborate to determine not only whether these technical hurdles can be overcome, but whether granting AI the autonomy to resolve hard choices is a desirable goal.

\section*{Acknowledgements}
Special thanks to Kevin Han Huang, David Abel, Cong Sun, and David Thorstad. I am grateful to Seth Lazar and audiences at the MINT Lab, to audiences at Google DeepMind, to Bohang Chen and audiences at Zhejiang University, and to Jiji Zhang and audiences at the Chinese University of Hong Kong, for discussions that substantially improved this paper. I also thank the anonymous reviewers, editors, and Area Chairs at FAccT 2026 and earlier review venues for feedback that sharpened the argument.

\section*{Competing Interests}
The author declares no competing interests.

\section*{Generative AI Usage Statement}
The author used Claude Opus 4.5/4.6 and Gemini 3/3.1 Pro for proofreading and grammar correction, language editing, formatting the reference list and citations, and general Overleaf usage questions. All arguments, analysis, and substantive content were written by the author without AI assistance.

\bibliographystyle{ACM-Reference-Format}
\bibliography{bibliography}

\appendix

\section{Ruth Chang and Bryce Goodman’s Alternative Diagnosis and Proposal}

Ruth Chang proposes that an agent faces a hard choice when external reasons given to them by the world ‘run out’ \cite{Chang2002, Chang2017, Chang2022, Chang2023, Chang2024b}. External reasons are grounded on values. They ‘run out’ when values grounding those reasons are in the same ‘neighbourhood’ but there is no fixed or precise exchange rate among them. This creates a unique kind of comparative relation between options which Chang calls ‘parity’. Chang distinguishes ‘parity’ from mere cases of indeterminacy by suggesting that in cases of ‘parity’, arbitrarily selecting an option is never intrinsically permissible. 

Chang’s framing is different from mine in several ways. First, my framing does not presuppose the nature of practical reason—external, desire-based, or will-based. I have my view on which theory of practical reason is suitable for understanding hard choices, but that is in another paper. Second, what I call cases of incommensurability or hard choices include but are not limited to what she calls cases of ‘parity’. I do not think that arbitrary picking is impermissible in all hard choices \cite{Tenenbaum2024}. Chang would say that there remains something pro tanto impermissible about flipping a coin in Lisbon vs Barcelona. I disagree.

However, here I do not want to discuss whether Chang’s theory is correct or whether her framing is the best. What matters here is Chang’s application of her theory on AI. Basically, Chang suggests that ‘current technological design (of AI) makes no room for hard choices’, yet since we have hard choices, suppose that ‘machines are to both align with our values and make choices for us, they should face hard choices too’, and therefore, ‘in hard choices, a machine should halt its processing and await human input’\cite{Chang2024a}. This application of her theory has severe problems.

First, Chang does not specify which sort of AI she is talking about—language models that can generate responses to our questions concerning choices that are hard to us, or algorithms that are directly used to make choices in areas such as robotics, auto-driving, healthcare, and finance. This ambiguity makes it difficult to assess the soundness of her impossibility conclusion. Chang does not provide more technical details to explain why her claim is true.

Bryce Goodman’s analysis \cite{Goodman2021}, which Chang cites, is more technically advised. However, Goodman discusses only supervised learning and reinforcement learning, leaving many other machine learning methods out of the sight and failing to reflect on the more general structure of optimisation. Moreover, Goodman’s discussion does not touch on LLMs (in his defence, Goodman published that paper before ChatGPT was launched). Although I eventually agree that current LLMs are unable to identify hard choices in the way I define, it is undeniable that if we ask the most advanced ones whether a choice that we find hard is hard and if yes why so, the likelihood is that the advanced LLM can actually say that it is hard and provide a comprehensive analysis on which option is better in regard to which value. Neither Chang nor Goodman addresses this important fact.

Second, Chang takes it for granted that if machines are to both align with our values and make choices for us, it is normatively necessary that they should be able to identify hard choices as we do \cite{Chang2024a}. However, Chang does not clearly specify why there is such a normative requirement in the first place. Earlier when clarifying what I am not suggesting in this paper, I have raised reasons to be sceptical of normative commitments like hers: First, we have always been fine with traditional non-AI decision-making computer algorithms that cannot identify hard choices either yet have long been used to make decisions for us. Second, in many cases, keeping a human in the loop will defeat the purpose of building automatic decision-making systems. I shall not repeat myself here.

Finally, since no AI agent or any computational algorithm responds to any practical reason, it is counterproductive to benchmark AIs against a definition of hard choice based on practical reason. Chang defines hard choices in terms of practical reason. Thus, it is counterproductive to apply her theory to AIs. 

Consider a choice between (A) getting to the destination a bit earlier with a 0.5 chance of killing a pedestrian and (B) getting to the destination a bit later while keeping all pedestrians safe. We can say that a human driver strictly prefers B to A and that is well in correspondence to the practical reasons they have. The driver is reasons-responsive. An auto-driving AI algorithm also chooses B instead of A every time, and this is robust against small improvements to B and A, probably because this well-aligned algorithm generates an ordering in which B is ranked over A. However, we cannot say that this AI algorithm is responding to practical reasons, because reasons-responsiveness cannot be reduced to a behavioural or modal property that can be trained or coded \cite{Heering2022, Safaei}, and one cannot respond to a practical reason without intentionality \cite{Kiesewetter2019}. And although what may be achieved in the future is anyone’s guess \cite{Long2024, GoldsteinKirkGiannini2025}, no known AI has intentionality. 

To study whether AIs can deal with hard choices, it is more productive to start with how AIs actually deal with any choice, which is my approach in this paper.

\end{document}